\documentclass[a4paper]{article}%
\usepackage{amsmath}%
\usepackage{amsfonts}%
\usepackage{amssymb}%
\usepackage{graphicx}
\usepackage{fullpage}

\usepackage{math_defs}

\begin{document}

\title{Unsupervised Adaptation of SPLDA}
\author{Jes\'{u}s Villalba\\\\
  Communications Technology Group (GTC),\\ Aragon Institute
  for Engineering Research (I3A),\\ University of Zaragoza, Spain\\
  \small \tt villalba@unizar.es}
\date{June 19, 2013}
\maketitle

\section{Introduction}
\label{sec:intro}

In this document we present a Variational Bayes solution to adapt a
SPLDA~\cite{villalba-splda} model to a new domain by using
unlabelled data. We assume that we count with a labelled dataset (for
example Switchboard) to initialise the model. 

\section{The Model}

\subsection{SPLDA}

SPLDA is a linear generative model where
an
i-vector $\phivec_{j}$ of speaker $i$ can be written as:
\begin{equation}
  \phivec_{j}=\muvec+\Vmat\yvec_{i}+\epsilon_{j}
  \label{eq:baysplda_model}
\end{equation} 
where $\muvec$ is a speaker independent mean, $\Vmat$ is the eigen-voices
matrix, $\yvec_i$ is the speaker factor vector, and $\epsilon$ is a channel
offset. 

We assume the following priors for $\yvec$ and $\epsilon$:
\begin{align}
  \label{eq:baysplda_yprior}
  \yvec_i&\sim\Gauss{\yvec_i}{\zerovec}{\Imat} \\
  \label{eq:baysplda_eprior}
  \epsilon_{j}&\sim\Gauss{\epsilon_{j}}{\zerovec}{\iWmat}
\end{align}
where $\mathcal{N}$ denotes a Gaussian distribution; $\Wmat$
is the within class precision matrix.

\begin{figure}[th]
  \begin{center}
    \includegraphics[width=0.60\textwidth]
    {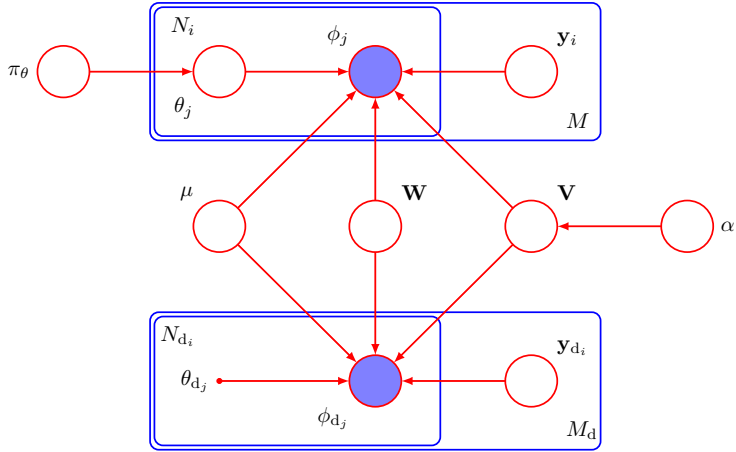}
  \end{center}
  \vspace{-0.5cm}
  \caption{BN for Bayesian SPLDA model.}
  \label{fig:bn_unsup_splda}
\end{figure}

Figure~\ref{fig:bn_unsup_splda} shows the case where the development
dataset is split into two parts: one part where the speaker labels
are known (supervised) and another with unknown labels (unsupervised).

We introduce the variables involved:
\begin{itemize}
\item Let $\Phimatd$ be the i-vectors of the supervised dataset.
\item Let $\Phimat$ be the i-vectors of the
  unsupervised dataset.
\item Let $\spk_i$ be the i-vectors belonging to the speaker $i$.
\item Let $\Ymatd$ be the speaker identity variables of the supervised
  dataset.
\item Let $\Ymat$ be the speaker identity variables of the unsupervised
  dataset.
\item Let $\thetad$ be the labelling of the supervised dataset. It
  partitions the $N_d$ i-vectors into $M_d$ speakers.
\item Let $\theta$ be the labelling of the unsupervised dataset. It
  partitions the $N$ i-vectors into $M$ speakers. $\theta_{j}$
  is a latent variable comprising a 1--of--M binary vector with
  elements $\theta_{ji}$ with $i=1,\dots,M$. This variable is
  equivalent to the cluster occupations of a GMM. The conditional
  distribution of $\theta$ given the weights of the mixture is:
  \begin{align}
    \Prob{\theta|\pi_{\theta}}=\prodjN
    \prodiM\pi_{\theta_{i}}^{\theta_{ji}} \;.
  \end{align}
\item Let $\pi_{\theta}$ be the weights of the mixture. We choose a
  Dirichlet prior for the weights:
  \begin{align}
    \Prob{\pi_\theta|\tau_0}=
    \mathrm{Dir}(\pi_\theta|\tau_0)=
    C(\tau_0)\prodiM \pi_{\theta_i}^{\tau_0-1}
  \end{align}
  where by symmetry we have chosen the same parameter $\tau_0$ for each of the
  components,
  and $C(\tau_0)$ is the normalisation constant for the Dirichlet
  distribution defined as
  \begin{align}
    C(\tau_0)=\frac{\Gamma(M\tau_0)}{\Gamma(\tau_0)^M}
  \end{align}
  and $\Gamma$ is the Gamma function.
\item Let $d$ be the i-vector dimension.
\item Let $n_y$ be the speaker factor dimension.
\item Let $\model=\left(\muvec,\Vmat,\Wmat\right)$ be the set of all
  the SPLDA parameters. In the most general case,
  we can assume that
  the parameters of the model are also
  hidden variables with prior and posterior distributions.
\end{itemize}

\subsection{Sufficient Statistics}
\label{sec:suff_stats}

We define some statistics for speaker $i$ in the unsupervised dataset:
\begin{align}
  N_i=&\sumjN \theta_{ji}\\
  \Fvec_i=&\sumjN \theta_{ji} \phivec_{j}\\
  \Smat_i=&\sumjNi \theta_{ji} \scatt{\phivec_{j}} \;.
\end{align}
We define the centered statistics as
\begin{align}
  \Fbar_i=&\Fvec_i-N_i \muvec\\
  \Sbarmat_i=&\sumjN \theta_{ji} \scattp{\phivec_{j}-\muvec}
  =\Smat_i-\muvec\Fvec_i^T-\Fvec_i\muvec^T+N_i\scatt{\muvec} \;.
\end{align}

We define the global statistics 
\begin{align}
  N&=\sumiM N_i \\
  \Fvec&=\sumiM \Fvec_i \\
  \Fbar&=\sumiM \Fbar_i \\
  \Smat&=\sumiM \Smat_i\\
  \Sbarmat&=\sumiM \Sbarmat_i \;.
\end{align}

Equally, we can define statistics for the supervised dataset:
$N_\rmd$, $\Fvec_{\rmd}$, $\Smat_{\rmd}$, etc

\subsection{Data conditional likelihood}
\label{sec:unsupsplda_cond}

The likelihood of the data given the hidden variables for speaker $i$
is 
\begin{align}
  \label{eq:unsupsplda_cond1}
  \lnProb{\Phimat_i|\yvec_i,\theta,\muvec,\Vmat,\Wmat}=&
  \sumjN \theta_{ji} 
  \ln\Gauss{\phivec_{j}}{\muvec+\Vmat\yvec_i}{\iWmat} \\
  \label{eq:unsupsplda_cond2}
  =&\frac{N_{i}}{2}\lndet{\frac{\Wmat}{2\pi}}
  -\med\sumjN \theta{ji} \mahP{\phivec_{j}}{\muvec-\Vmat\yvec_i}{\Wmat}\\
  \label{eq:unsupsplda_cond3}
  =&\frac{N_{i}}{2}\lndet{\frac{\Wmat}{2\pi}}
  -\med\trace\left(\Wmat\Sbarmat_{i}\right)
  +\yvec_i^{T}\Vmat^{T}\Wmat\Fbar_{i}
  -\frac{N_{i}}{2}\yvec_i^T\Vmat^T\Wmat\Vmat\yvec_i\;.
\end{align}

We can also write this likelihood as:
\begin{align}
  \label{eq:unsupsplda_cond4}
  \lnProb{\spk_i|\yvec_i,\theta,\muvec,\Vmat,\Wmat}=&
  \frac{N_{i}}{2}\lndet{\frac{\Wmat}{2\pi}}
  -\med\trace\left(\Wmat\left(\Smat_{i}
      -2\Fvec_{i}\muvec^{T}
      +N_{i}\scatt{\muvec}
    \right.\right.\\
  &\left.\left.-2 \left(\Fvec_{i}-N_{i}\muvec\right)\yvec_i^{T}\Vmat^{T}
      +N_{i} \Vmat\yvec_i\yvec_i^T\Vmat^T\right)\right)\;.
\end{align}

If we define:
\begin{align}
  \ytildevec_i=
  \begin{bmatrix}
    \yvec_i\\
    1
    \end{bmatrix}
    , & \quad \Vtildemat=
  \begin{bmatrix}
    \Vmat & \muvec
  \end{bmatrix}
\end{align}
we can write it as
\begin{align}
  \label{eq:baysplda_cond5}
  \lnProb{\spk_i|\yvec_i,\theta,\muvec,\Vmat,\Wmat}=&
  \sumjNi \theta_{ji} \ln
  \Gauss{\phivec_{ij}}{\Vtildemat\ytildevec_i}{\iWmat} \\
  \label{eq:baysplda_cond8}
  =&\frac{N_i}{2}\lndet{\frac{\Wmat}{2\pi}}
  -\med\trace\left(\Wmat\left(\Smat_i-2\Fvec_i\ytildevec_i^{T}\Vtildemat^{T}
     +N_i \Vtildemat\ytildevec_i\ytildevec_i^T\Vtildemat^T\right)\right)\;.
\end{align}

\section{Variational Inference with Point Estimates of $\muvec$,
  $\Vmat$ and $\Wmat$}
\label{sec:unsupsplda_v1}

\begin{figure}[th]
  \begin{center}
    \includegraphics[width=0.60\textwidth]
    {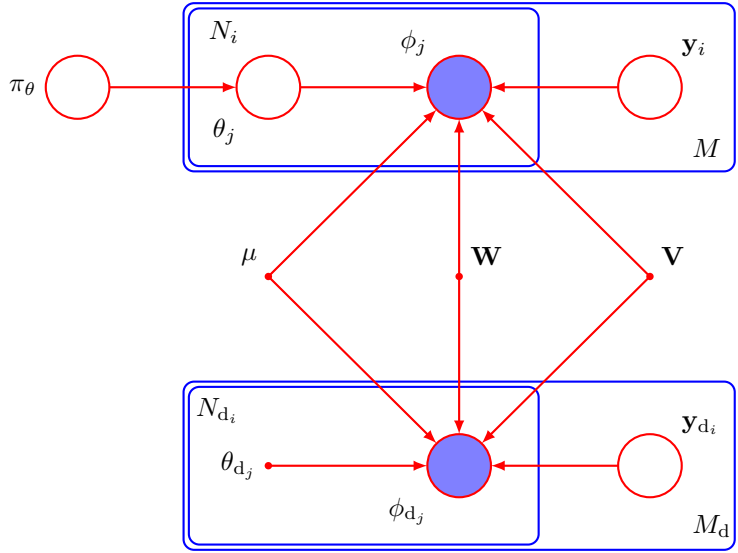}
  \end{center}
  \vspace{-0.5cm}
  \caption{BN for SPLDA with point estimates of the model parameters.}
  \label{fig:bn_unsup_splda_pe}
\end{figure}

As first approximation, we assume a simplified model where we take
point estimates of the parameters $\muvec$, $\Vmat$ and $\Wmat$. In
this case, the graphical model simplifies to the one in
Figure~\ref{fig:bn_unsup_splda_pe}.

In this model, $\yvec_i$, $\yvec_{\rmd i}$ and $\theta_{ij}$ are
the only
hidden variables. $\Vmat$, $\muvec$ and $\Wmat$ are hyperparameters
that can be obtained by maximising the VB lower bound. 

\subsection{Variational Distributions}
\label{sec:unsupsplda_v1_q}

We write the joint distribution of the observed and latent factors:
\begin{align}
  \Prob{\Phimat,\Phimatd,\Ymat,\Ymatd,\theta,\pi_{\theta}|
    \thetad,\tau_0,\muvec,\Vmat,\Wmat}=&
  \Prob{\Phimat|\Ymat,\theta,\muvec,\Vmat,\Wmat}\Prob{\Ymat} 
  \Prob{\theta|\pi_{\theta}}\Prob{\pi_{\theta}|\tau_0} \nonumber\\
  &\Prob{\Phimatd|\Ymatd,\thetad,\muvec,\Vmat,\Wmat}\Prob{\Ymatd} \;.
\end{align}
Following, the conditioning on
$\left(\thetad,\tau_0,\muvec,\Vmat,\Wmat\right)$ will be dropped for
convenience. 

Now, we consider the partition of the posterior:
\begin{align}
  \Prob{\Ymat,\Ymatd,\theta,\pi_\theta|\Phimat,\Phimatd}\approx
  \q{\Ymat,\Ymatd,\theta}=
  \q{\Ymat,\Ymatd}\q{\theta}\q{\pi_\theta} \;.
\end{align}

The optimum for $\qopt{\Ymat,\Ymatd}$:
\begin{align}
  \lnqopt{\Ymat,\Ymatd}=&
  \Expcond{\lnProb{\Phimat,\Phimatd,\Ymat,\Ymatd,\theta,\pi_{\theta}}}
    {\theta,\pi_{\theta}}+\const\\
    =&\Expcond{\lnProb{\Phimat|\Ymat,\theta}}{\theta}
    +\lnProb{\Ymat}
    +\lnProb{\Phimatd|\Ymatd}
    +\lnProb{\Ymatd}
    +\const\\
    =&\sumiM \yvec_i^T\Vmat^T\Wmat\Exp{\Fbar_i} 
    -\med
    \yvec_i^T\left(\Imat+\Exp{N_i}\Vmat^T\Wmat\Vmat\right)\yvec_i
    \nonumber \\
    &\sumiMd \yvec_{\rmd_i}^T\Vmat^T\Wmat\Fbar_{\rmd i} 
    -\med \yvec_{\rmd_i}^T\left(\Imat+N_{\rmd_i}\Vmat^T\Wmat\Vmat\right)\yvec_{\rmd_i}
    +\const
\end{align}
Therefore $\qopt{\Ymat,\Ymatd}$ is a product of Gaussian distributions.
\begin{align}
   \qopt{\Ymat,\Ymatd}=&\prodiM \Gauss{\yvec_i}{\ybarvec_i}{\iLmatyi}
   \prodiMd \Gauss{\yvec_{\rmd_i}}{\ybarvec_{\rmd_i}}{\iLmatydi}\\
   \Lmatyi=&\Imat+\Exp{N_i}\Vmat^T\Wmat\Vmat\\
   \ybarvec_i=&\iLmatyi\Vmat^T\Wmat\Exp{\Fbar_i}\\
   \label{eq:unsupsplda_v1_ENi}
   \Exp{N_i}=&\sumjN \Exp{\theta_{ji}}\\
   \label{eq:unsupsplda_v1_EFi}
   \Exp{\Fbar_i}=&\sumjN \Exp{\theta_{ji}}\left(\phi_j-\muvec\right)\\
   \Lmatydi=&\Imat+N_{\rmd_i}\Vmat^T\Wmat\Vmat\\
   \ybarvec_{\rmd_i}=&\iLmatydi\Vmat^T\Wmat\Fbar_{\rmd_i}
\end{align}

The optimum for $\qopt{\theta}$:
\begin{align}
  \lnqopt{\theta}=&
  \Expcond{\lnProb{\Phimat,\Phimatd,\Ymat,\Ymatd,\theta,\pi_{\theta}}}
  {\Ymat,\Ymatd,\pi_\theta}\\
  =&\Expcond{\lnProb{\Phimat|\Ymat,\theta}}{\Ymat}
  +\Expcond{\lnProb{\theta|\pi_\theta}}{\pi_\theta}+\const\\
  =&\sumjN\sumiM \theta_{ji} \left[ \med\lndet{\frac{\Wmat}{2\pi}}
  -\med\Exp{\mahP{\phivec_{j}}{\muvec-\Vmat\yvec_i}{\Wmat}}
  +\Exp{\ln\pi_{\theta_i}}\right]+\const\\
  =&\sumjN\sumiM \theta_{ji} \left[ \med\lndet{\frac{\Wmat}{2\pi}}
    -\med\mahP{\phivec_{j}}{\muvec}{\Wmat}
    +\Exp{\yvec_i}^T\Vmat^T \Wmat\left(\phi_j-\muvec\right)
  \right.\nonumber\\
  &\left.\quad\quad\quad\quad
    -\med\trace\left(\Vmat^T\Wmat\Vmat\Exp{\scatt{\yvec_i}}\right)
    +\Exp{\ln\pi_{\theta_i}}\right]+\const\;.
\end{align}

Taking exponentials in both sides:
\begin{align}
\qopt{\theta}=&\prodjN\prodiM r_{ji}^{\theta_{ji}}
\end{align}
where
\begin{align}
  r_{ji}=&\frac{\varrho_{ji}}{\sumiM \varrho_{ji}}\\
  \ln \varrho_{ji}=&
  \med\lndet{\frac{\Wmat}{2\pi}}
  -\med\mahP{\phivec_{j}}{\muvec}{\Wmat}
    +\Exp{\yvec_i}^T\Vmat^T \Wmat\left(\phi_j-\muvec\right)\\
    &-\med\trace\left(\Vmat^T\Wmat\Vmat\Exp{\scatt{\yvec_i}}\right)
    +\Exp{\ln\pi_{\theta_i}}\;
\end{align}

The optimum for $\qopt{\pi_\theta}$:
\begin{align}
  \lnqopt{\pi_\theta}=&
  \Expcond{\lnProb{\Phimat,\Phimatd,\Ymat,\Ymatd,\theta,\pi_{\theta}}}
  {\Ymat,\Ymatd,\theta}\\
  =&\Expcond{\lnProb{\theta|\pi_\theta}}{\theta}
  +\lnProb{\pi_\theta|\tau_0}+\const\\
  =&\sumjN\sumiM\Exp{\theta_{ji}}\ln\pi_{\theta_i}
  +(\tau_0-1)\sumiM\ln\pi_\theta+\const\\
  =&\sumiM \left(\Exp{N_i}+\tau_0-1\right)\ln\pi_{\theta_i} \;.
\end{align}
Thus:
\begin{align}
\qopt{\pi_\theta}=&\mathrm{Dir}(\pi_\theta|\tau)=
C(\tau)\prodiM \pi_{\theta_i}^{\tau_i-1}
\end{align}
where
\begin{align}
  \tau_i=&\Exp{N_i}+\tau_0\\
  C(\tau)=&\frac{\gammaf{\sumiM\tau_i}}{\prodiM\gammaf{\tau_i}}\;
\end{align}

Finally, we evaluate the expectations:
\begin{align}
  \Exp{\yvec_i}=&\ybarvec_i\\
  \Exp{\scatt{\yvec_i}}=&\iLmatyi+\scatt{\ybarvec_i}\\
  \Exp{\scatt{\ytildevec_i}}=&
  \begin{bmatrix}
    \Exp{\scatt{\yvec_i}} & \Exp{\yvec_i} \\
    \Exp{\yvec_i}^T & 1 
  \end{bmatrix}\\
\Exp{\theta_{ji}}=&r_{ji}\\
\Exp{\pi_{\theta_i}}=&\frac{\tau_i}{\sumiM\tau_i}\\
\Exp{\ln \pi_{\theta_i}}=&\psi\left(\tau_i\right)
-\psi\left(\sumiM\tau_i\right)
\end{align}

\subsubsection{Distributions with deterministic annealing}

If we use annealing, for a parameter $\kappa$, we have
\begin{align}
   \qopt{\Ymat,\Ymatd}=&\prodiM \Gauss{\yvec_i}{\ybarvec_i}{1/\kappa\;\iLmatyi}
   \prodiMd \Gauss{\yvec_{\rmd_i}}{\ybarvec_{\rmd_i}}{1/\kappa\;
     \iLmatydi}\;
\end{align}
\begin{align}
\qopt{\theta}=&\prodjN\prodiM r_{ji}^{\theta_{ji}}
\end{align}
where
\begin{align}
  r_{ji}=&\frac{\varrho_{ji}^\kappa}{\sumiM \varrho_{ji}^\kappa}\;
\end{align}
\begin{align}
\qopt{\pi_\theta}=&\mathrm{Dir}(\pi_\theta|\tau)=
C(\tau)\prodiM \pi_{\theta_i}^{\tau_i-1}
\end{align}
where
\begin{align}
  \tau_i=&\kappa(\Exp{N_i}+\tau_0-1)+1
\end{align}

\subsection{Variational lower bound}
\label{sec:unsupsplda_v1_lb}

The lower bound is given by:
\begin{align}
\lowb=&\Expcond{\lnProb{\Phimat|\Ymat,\theta}}{\Ymat,\theta}
+\Expcond{\lnProb{\Ymat}}{\Ymat} 
+\Expcond{\lnProb{\theta|\pi_{\theta}}}{\theta,\pi_\theta}
+\Expcond{\lnProb{\pi_{\theta}}}{\pi_{\theta}} \nonumber\\
  &+\Expcond{\lnProb{\Phimatd|\Ymatd}}{\Ymatd}
  +\Expcond{\lnProb{\Ymatd}}{\Ymatd} \nonumber\\
  &-\Expcond{\lnq{\Ymat}}{\Ymat}-\Expcond{\lnq{\theta}}{\theta}
  -\Expcond{\lnq{\pi_\theta}}{\pi_\theta}
  -\Expcond{\lnq{\Ymatd}}{\Ymatd}\;.
\end{align}

The term $\Expcond{\lnProb{\Phimat|\Ymat,\theta}}{\Ymat,\theta}$:
\begin{align}
  \Expcond{\lnProb{\Phimat|\Ymat,\theta}}{\Ymat,\theta}=&
  \frac{\Exp{N}}{2}\lndet{\frac{\Wmat}{2\pi}}\nonumber\\
  &-\med\trace\left(\Wmat\left(\Exp{\Smat}
      -2\sumiM\Exp{\Fvec_i}\Exp{\ytildevec_i}^{T}\Vtildemat^{T}
      +\Vtildemat\sumiM \Exp{N_i}\Exp{\scatt{\ytildevec_i}}\Vtildemat^T
    \right)\right)
\end{align}
We define
\begin{align}
  \Cmatytilde=&\sumiM\Exp{\Fvec_i}\Exp{\ytildevec_i}^{T}\\
  \Rmatytilde=&\sumiM \Exp{N_i}\Exp{\scatt{\ytildevec_i}}
\end{align}

Then
\begin{align}
  \Expcond{\lnProb{\Phimat|\Ymat,\theta}}{\Ymat,\theta}=&
  \frac{\Exp{N}}{2}\lndet{\frac{\Wmat}{2\pi}}
  -\med\trace\left(\Wmat\left(\Exp{\Smat}
      -2\Cmatytilde\Vtildemat^{T}
      +\Vtildemat\Rmatytilde\Vtildemat^T
    \right)\right)\;.
\end{align}

The term $\Expcond{\lnProb{\Phimatd|\Ymatd}}{\Ymatd}$:
\begin{align}
  \Expcond{\lnProb{\Phimatd|\Ymatd}}{\Ymatd}=&
  \frac{N_\rmd}{2}\lndet{\frac{\Wmat}{2\pi}}
  -\med\trace\left(\Wmat\left(\Smat_\rmd
      -2\Cmatytilded\Vtildemat^{T}
      +\Vtildemat\Rmatytilded\Vtildemat^T
    \right)\right)\;.
\end{align}
where
\begin{align}
  \Cmatytilded=&\sumiMd \Fvec_{\rmd_i}\Exp{\ytildevec_{\rmd_i}}^{T}\\
  \Rmatytilded=&\sumiMd N_{\rmd_i}\Exp{\scatt{\ytildevec_{\rmd_i}}}
\end{align}

The term $\Expcond{\lnProb{\Ymat}}{\Ymat}$:
\begin{align}
  \Expcond{\lnProb{\Ymat}}{\Ymat}=&
  -\frac{M n_y}{2}\ln(2\pi)-\med\trace\left(\Rhomaty\right)
\end{align}
where
\begin{align}
  \Rhomat=\sumiM \Exp{\scatt{\yvec_i}}
\end{align}

The term $\Expcond{\lnProb{\Ymatd}}{\Ymatd}$:
\begin{align}
  \Expcond{\lnProb{\Ymatd}}{\Ymatd}=&
  -\frac{\Md n_y}{2}\ln(2\pi)-\med\trace\left(\Rhomatyd\right)
\end{align}
where
\begin{align}
  \Rhomatyd=\sumiMd \Exp{\scatt{\yvec_{\rmd_i}}}
\end{align}

The term $\Expcond{\lnProb{\theta|\pi_{\theta}}}{\theta,\pi_\theta}$:
\begin{align}
\Expcond{\lnProb{\theta|\pi_{\theta}}}{\theta,\pi_\theta}=&
\sumjN\sumiM r_{ji} \Exp{\ln\pi_{\theta_i}}
\end{align}

The term $\Expcond{\lnProb{\pi_{\theta}}}{\pi_{\theta}}$:
\begin{align}
\Expcond{\lnProb{\pi_{\theta}}}{\pi_{\theta}}=&
\ln C(\tau_0)+(\tau_0-1)\sumiM \Exp{\ln\pi_{\theta_i}}
\end{align} 

The term $\Expcond{\lnq{\Ymat}}{\Ymat}$:
\begin{align}
  \Expcond{\lnq{\Ymat}}{\Ymat}=&
  -\frac{Mn_y}{2}(\ln(2\pi)+1)+\med\sumiM\lndet{\Lmatyi}
\end{align}

The term $\Expcond{\lnq{\Ymatd}}{\Ymatd}$:
\begin{align}
  \Expcond{\lnq{\Ymatd}}{\Ymatd}=&
  -\frac{\Md n_y}{2}(\ln(2\pi)+1)+\med\sumiMd\lndet{\Lmatydi}
\end{align}

The term $\Expcond{\lnq{\theta}}{\theta}$:
\begin{align}
\Expcond{\lnq{\theta}}{\theta}=&\sumjN\sumiM r_{ji}\ln r_{ji}
\end{align}

The term $\Expcond{\lnq{\pi_\theta}}{\pi_\theta}$:
\begin{align}
\Expcond{\lnq{\pi_\theta}}{\pi_\theta}=&
\ln C(\tau) + \sumiM (\tau_i-1) \Exp{\ln \pi_{\theta_i}}
\end{align}

\subsection{Hyperparameter optimisation}
\label{sec:unsupsplda_v1_hyp}

We can obtain the hyperparameters $(\tau_0,\muvec,\Vmat,\Wmat)$ by maximising
the lower bound. We control the weight of each of the databases on the
estimation by introducing the parameter $\eta\leq1$ into the lower
bound expression:
\begin{align}
  \lowb(\muvec,\Vmat,\Wmat,\tau_0)=&
  \Expcond{\lnProb{\Phimat|\Ymat,\theta}}{\Ymat,\theta}
  +\Expcond{\lnProb{\pi_{\theta}}}{\pi_{\theta}}
  +\eta\Expcond{\lnProb{\Phimatd|\Ymatd}}{\Ymatd}+\const
\end{align}

We derive for $\Vtildemat$:
\begin{align}
  \frac{\partial\lowb}{\partial \Vtildemat}=&
  \Cmatytilde+\eta\Cmatytilded
  -\Vtildemat\left(\Rmatytilde+\eta\Rmatytilded\right)=\zerovec 
  \implies\\
  &\Vtildemat=\Cmatytilde^{\prime}
  \Rmatytilde^{\prime -1}
\end{align}
where
\begin{align}
\Cmatytilde^{\prime}=&\Cmatytilde+\eta\Cmatytilded\\
\Rmatytilde^{\prime}=&\Rmatytilde+\eta\Rmatytilded
\end{align}

We derive for $\Wmat$:
\begin{align}
  \frac{\partial \lowb}{\partial \Wmat}=&
  \frac{\Exp{N}+\eta N_\rmd}{2}
  \left(2\iWmat-\diag\left(\Wmat^{-1}\right)\right)
-\med\left(\Kmat+\Kmat^T-\diag\left(\Kmat\right)\right)
\end{align}
where 
\begin{align}
  \Kmat=\Exp{\Smat}+\eta\Smat_{\rmd}
  -2\Cmatytilde^{\prime}\Vtildemat
  +\Vtildemat\Rmatytilde^{\prime}\Vtildemat^T
\end{align}
Then
\begin{align}
  \iWmat=\frac{1}{\Exp{N}+\eta N_\rmd}\frac{\Kmat+\Kmat^T}{2}
\end{align}

We derive for $\tau_0$:
\begin{align}
  \frac{\partial \lowb}{\partial \tau_0}=&
  M\left(\psi\left(M\tau_0\right)-\psi\left(\tau_0\right)\right)
  +\sumiM \Exp{\ln \pi_{\theta_i}}=0
\end{align}

We define
$\tau_0=\exp(\tilde{\tau}_0)$ and
\begin{align}
  f(\tau_0)=&\psi\left(M\tau_0\right)-\psi\left(\tau_0\right)+g=0\\
  g=&\frac{1}{M}\sumiM \Exp{\ln \pi_{\theta_i}}=0\;.
\end{align}
We can solve for $\tilde{\tau}_0$ by Newton-Rhapson iterations:
\begin{align}
  \tilde{\tau}_{0_{new}}=&\tilde{\tau}_0
  -\frac{f(\tilde{\tau}_0)}{f^\prime(\tilde{\tau}_0)}\\
  =&\tilde{\tau}_0
  -\frac{\psi\left(M\tau_0\right)
      -\psi\left(\tau_0\right)+g}
  {\tau_0\left(\psi^\prime\left(M\tau_0\right)
      -\psi^\prime\left(\tau_0\right)\right)}
\end{align}
Taking exponentials in both sides:
\begin{align}
  \tau_{0_{new}}=&\tau_0 \exp\left(
  -\frac{\psi\left(M\tau_0\right)
      -\psi\left(\tau_0\right)+g}
  {\tau_0\left(\psi^\prime\left(M\tau_0\right)
      -\psi^\prime\left(\tau_0\right)\right)}\right)
\end{align}

\subsection{Minimum divergence}
\label{sec:unsupsplda_v1_md}

We assume a more general prior for the hidden variables:
\begin{align}
  \Prob{\yvec}=\Gauss{\yvec}{\muvecy}{\iLambmaty}
\end{align}

Then we maximise
\begin{align}
  \lowb(\muvecy,\Lambmaty)=&
  \sumiM \ExpcondY{\ln\Gauss{\yvec}{\muvecy}{\iLambmaty}}
  +\eta \sumiMd \ExpcondY{\ln\Gauss{\yvec_\rmd}{\muvecy}{\iLambmaty}}\\
  =&\frac{M+\eta \Md}{2}\lndet{\Lambmaty} \nonumber\\
  &-\med\trace\left(\Lambmaty \left(
      \sumiM \Exp{\scattp{\yvec_i-\muvecy}}
      + \eta \sumiMd \Exp{\scattp{\yvec_{\rmd_i}-\muvecy}}
    \right)\right)+\const
\end{align}

We derive for $\muvecy$:

\begin{align}
\frac{\partial\lowb(\muvecy,\Lambmaty)}{\partial\muvecy}&=
\med\sumiM\Lambmaty\Exp{\yvec_i-\muvecy}
+\eta \med\sumiMd\Lambmaty\Exp{\yvec_{\rmd_i}-\muvecy}
=\zerovec \quad \implies\\
\muvecy&=\frac{1}{M+\eta\Md}
\left(\sumiM\Exp{\yvec_i}+\eta \sumiMd\Exp{\yvec_{\rmd_i}}\right)
\end{align}

We derive for $\Lambmaty$:
\begin{align}
\frac{\partial\lowb(\muvecy,\Lambmaty)}{\partial\Lambmaty}&=
\frac{M+\eta\Md}{2}\left(2\Lambmaty^{-1}-\mathrm{diag}(\Lambmaty^{-1})\right)
-\med\left(2\Smat-\mathrm{diag}(\Smat)\right)=\zerovec
\end{align}
where 
\begin{align}
  \Smat=\sumiM\Exp{\scattp{\yvec_i-\muvecy}}+
  \eta \sumiM\Exp{\scattp{\yvec_{\rmd_i}-\muvecy}}
\end{align}
Then

\begin{align}
  \Sigmaty=\Lambmaty^{-1}=
  \frac{1}{M+\eta\Md}\left(\Rhomaty+\eta\Rhomatyd\right)-\scatt{\muvecy}
\end{align}

To obtain a standard prior for $\yvec$
We transform $\muvec$ and $\Vmat$ by using
\begin{align}
  \muvec^\prime=&\muvec+\Vmat\muvecy\\
  \Vmat^\prime=&\Vmat(\Sigmaty^{1/2})^T
\end{align}

\subsection{Determining the number of speakers}
\label{sec:unsupsplda_v1_nspk}

To determine the number of speakers we initialise the algorithm
assuming that there is a large number of speakers and after some
iterations we eliminate speakers based on heuristics:
\begin{itemize}
  \item Each i-vector belongs only to one speaker.
  \item Each speaker has an integer number of i-vectors.
  \item If several i-vectors have similar $\Exp{\theta}$ for several
    speakers we can merge the speakers.
  \item Compare the lower bound for different values of $M$ to
    determine the best number of speakers.
\end{itemize}

\subsection{Initialise the VB}
\label{sec:unsupsplda_v1_init}

\begin{itemize}
\item The values of $\muvec$, $\Vmat$ and $\Wmat$ can be initialised using
  the supervised dataset.
\item $\q{\pi_\theta}$ can be initialised assuming that all the speakers
  have the same number of i-vectors.
\item $\q{\theta}$ can be initialised using AHC or some simple
  algorithm based on the pairwise scores computed evaluating the initial
  PLDA model. We should also initialise $\q{\theta}$ with the oracle labels
  and check that the partition does not degrade itself as the
  algorithm iterates. This will provide an upper bound for the
  performance of the algorithm.
\item Instead of initialising $\q{\theta}$ we can initialise
  $\q{\Ymat}$ sampling random speakers from the standard
  distribution and afterwards, compute $\q{\theta}$ given $\q{\Ymat}$.
\end{itemize}

\subsection{Combining VB and sampling methods}

I am interested in Dan's idea of combining VB and sampling
methods. Instead of computing the i-vector statistics as shown in
Equations~\eqref{eq:unsupsplda_v1_ENi}
and~\eqref{eq:unsupsplda_v1_EFi}, we can 
draw samples $\hat{\theta}_{jk}$,  $k=1,\dots,K$ from
$\q{\theta}$. Then, compute $K$ i-vector statistics for speaker $i$ as:
\begin{align}
  &N_{ik}=\sumjN \hat{\theta}_{jki} &
  \Fvec_{ik}=\sumjN \hat{\theta}_{jki}\phivec_j\;.
\end{align} 
Thus, the statistics are computed in a way that each i-vector only
belongs to one speaker while in the standard VB formulation i-vectors
are shared between several clusters.
Then, we can follow several strategies:
\begin{itemize}
  \item Select the sample $k^{*}$ that maximises the lower bound.
  \item For sample $k$, obtain the accumulators needed to compute
    $\muvec$, $\Vmat$ and $\Wmat$ ($\Rmatytilde$, $\Cmatytilde$, etc), 
    average the accumulators of all the
    samples and compute the model.
  \item For each sample $k$, compute a model and average the
    models. However, I think that averaging the accumulators is more
    correct. 
\end{itemize}
The drawback of this method is that the computational cost grows linearly
with $K$, and we may need a large $K$ to make it work.

\section{Variational inference with Gaussian-Gamma priors for $\Vmat$,
  Gaussian for $\muvec$ and non-informative prior for $\Wmat$}
\label{sec:unsupsplda_v2}

\subsection{Model priors}
\label{sec:unsupsplda_v2_priors}

We chose the model priors based on the Bishop's paper about VB
PPCA~\cite{Bishop1999}. 
We introduce a \emph{hierarchical} prior $\Prob{\Vmat|\alphavec}$ over
the matrix $\Vmat$ governed by a $n_y$ dimensional vector of
hyperparameters where $n_y$ is the dimension of the factors. Each
hyperparameter controls one of the columns of the matrix $\Vmat$
through a conditional Gaussian distribution of the form:
\begin{align}
  \label{eq:unsupsplda_v2_priors_V}
  \Prob{\Vmat|\alphavec}=
  \prod_{q=1}^{n_y}\left(\frac{\alpha_q}{2\pi}\right)^{d/2}
  \exp\left(-\med\alpha_q\vvec_q^T\vvec_q\right)
\end{align}
where $\vvec_q$ are the columns of $\Vmat$. Each $\alpha_q$ controls
the inverse variance of the corresponding $\vvec_q$. If a particular
$\alpha_q$ has a posterior distribution concentrated at large
values, the corresponding $\vvec_q$ will tend to be small, and that
direction of the latent space will be effectively 'switched off'. 

We define a prior for $\alphavec$:
\begin{align}
  \Prob{\alphavec}=
  \prod_{q=1}^{n_y}\Gammad{\alpha_q}{a_{\alpha}}{b_{\alpha}}
\end{align}
where $\mathcal{G}$ denotes the Gamma distribution. Bishop defines broad
priors setting $a=b=10^{-3}$. 

We place a Gaussian prior for the mean $\muvec$:
\begin{align}
  \Prob{\muvec}=\Gauss{\muvec}{\muvec_0}{\diag(\betavec)^{-1}}\;.
\end{align}
We will consider the case where each dimension has
different precision and the case with isotropic precision 
($\diag(\betavec)=\beta\Imat$). 

Finally, we use a non-informative prior for $\Wmat$ like
in~\cite{villalba-bay2cov}. 
\begin{align}
\label{eq:jpriorb1}
\Prob{\Wmat}&=\lim_{k \to 0}\Wishart{\Wmat}{\Wmat_0/k}{k}\\
\label{eq:jpriorb2}
&=\alpha \left|\Wmat\right|^{-(d+1)/2}\;.
\end{align}

\subsection{Variational distributions}
\label{sec:unsupsplda_v2_q}

We write the joint distribution of the observed and latent variables:
\begin{align}
  \Prob{\Phimat,\Phimatd,\Ymat,\Ymatd,\theta,
    \pi_\theta,\muvec,\Vmat,\Wmat,\alphavec
    |\theta_\rmd,\tau_0,\muvec_0,\betavec,a_{\alpha},b_{\alpha}}=&
  \Prob{\Phimat|\Ymat,\theta,\muvec,\Vmat,\Wmat}\Prob{\Ymat}
  \Prob{\theta|\pi_{\theta}}\Prob{\pi_{\theta}|\tau_0} \nonumber\\
  &\Prob{\Vmat|\alphavec}\Prob{\alphavec|a,b}
  \Prob{\muvec|\muvec_0,\beta}\Prob{\Wmat} \nonumber\\
  &\Prob{\Phimatd|\Ymatd,\theta_\rmd,\muvec,\Vmat,\Wmat}\Prob{\Ymatd}
\end{align}
Following, the conditioning on
$\left(\theta_\rmd,\tau_0,\muvec_0,\betavec,a_{\alpha},b_{\alpha}\right)$ 
will be dropped for convenience. 

Now, we consider the partition of the posterior:
\begin{align}
  \Prob{\Ymat,\Ymatd,\theta,\pi_\theta,\muvec,\Vmat,\Wmat,\alphavec
    |\Phimat,\Phimatd}\approx&
  \q{\Ymat,\Ymatd,\theta,\pi_\theta,\muvec,\Vmat,\Wmat,\alphavec}
\nonumber\\
=&\q{\Ymat,\Ymatd}\q{\theta}\q{\pi_\theta}
  \prod_{r=1}^d\q{\vtildevec'_r}\q{\Wmat}\q{\alphavec}
\end{align}
where $\vtildevec'_r$ is a column vector containing the $r^{th}$ row of
$\Vtildemat$. If $\Wmat$ were a diagonal matrix the factorisation 
$\prod_{r=1}^d\q{\vtildevec'_r}$ is not necessary because it arises
naturally when solving the posterior. However, for full covariance
$\Wmat$, the posterior of $\vec(\Vtildemat)$ is a Gaussian
with a huge full covariance matrix. We force the factorisation
to made the problem tractable. 

The optimum for $\qopt{\Ymat,\Ymatd}$:
\begin{align}
  \lnqopt{\Ymat,\Ymatd}=&
  \Expcond{\lnProb{\Phimat,\Phimatd,\Ymat,\Ymatd,\theta,\pi_\theta,
      \muvec,\Vmat,\Wmat,\alphavec}}
  {\theta,\pi_\theta,\muvec,\Vmat,\Wmat,\alphavec}+\const\\
    =&\Expcond{\lnProb{\Phimat|\Ymat,\theta,\muvec,\Vmat,\Wmat}}
    {\theta,\muvec,\Vmat,\Wmat}
    +\lnProb{\Ymat}\nonumber\\
    &+\Expcond{\lnProb{\Phimatd|\Ymatd,\muvec,\Vmat,\Wmat}}
    {\muvec,\Vmat,\Wmat}
    +\lnProb{\Ymatd}
    +\const\\
    =&\sumiM \yvec_i^T\Exp{\Vmat^T\Wmat\left(\Fvec_i-N_i\muvec\right)} 
    -\med
    \yvec_i^T\left(\Imat+\Exp{N_i}\Exp{\Vmat^T\Wmat\Vmat}\right)\yvec_i
    \nonumber\\
    &+\sumiMd \yvec_{\rmd_i}^T
    \Exp{\Vmat^T\Wmat\left(\Fvec_{\rmd_i}-N_{\rmd_i}\muvec\right)} 
    -\med
    \yvec_{\rmd_i}^T
    \left(\Imat+N_{\rmd_i}\Exp{\Vmat^T\Wmat\Vmat}\right)
    \yvec_{\rmd_i}
    +\const
\end{align}
Therefore $\qopt{\Ymat,\Ymatd}$ is a product of Gaussian distributions.
\begin{align}
  \qopt{\Ymat,\Ymatd}=&
  \prodiM \Gauss{\yvec_i}{\ybarvec_i}{\iLmatyi}
   \prodiMd \Gauss{\yvec_{\rmd_i}}{\ybarvec_{\rmd_i}}{\iLmatydi}\\
   \Lmatyi=&\Imat+\Exp{N_i}\Exp{\Vmat^T\Wmat\Vmat}\\
   \ybarvec_i=&\iLmatyi
   \left(\Exp{\Vmat}^T\Exp{\Wmat}\Exp{\Fvec_i}
     -\Exp{N_i}\Exp{\Vmat^T\Wmat\muvec}\right)\\
   \Exp{N_i}=&\sumjN \Exp{\theta_{ji}}\\
   \Exp{\Fvec_i}=&\sumjN \Exp{\theta_{ji}}\phi_j\\
   \Lmatydi=&\Imat+N_{\rmd_i}\Exp{\Vmat^T\Wmat\Vmat}\\
   \ybarvec_{\rmd_i}=&\iLmatydi
   \left(\Exp{\Vmat}^T\Exp{\Wmat}\Fvec_{\rmd_i}
     -N_{\rmd_i}\Exp{\Vmat^T\Wmat\muvec}\right)
\end{align}

The optimum for $\qopt{\theta}$:
\begin{align}
  \lnqopt{\theta}=&
  \Expcond{\lnProb{\Phimat,\Phimatd,\Ymat,\Ymatd,\theta,\pi_\theta,
      \muvec,\Vmat,\Wmat,\alphavec}}
  {\Ymat,\Ymatd,\pi_\theta,\muvec,\Vmat,\Wmat,\alphavec}+\const\\
  =&\Expcond{\lnProb{\Phimat|\Ymat,\theta,\muvec,\Vmat,\Wmat}}
  {\Ymat,\muvec,\Vmat,\Wmat}
  +\Expcond{\lnProb{\theta|\pi_\theta}}{\pi_\theta}+\const\\
  =&\sumjN\sumiM \theta_{ji} \left[ 
    \med\Exp{\lndet{\Wmat}}-\frac{d}{2}\ln(2\pi)
  -\med\Exp{\mahP{\phivec_{j}}{\Vtildemat\ytildevec_i}{\Wmat}}
  +\Exp{\ln\pi_{\theta_i}}\right]+\const
\end{align}

Taking exponentials in both sides:
\begin{align}
\qopt{\theta}=&\prodjN\prodiM r_{ji}^{\theta_{ji}}
\end{align}
where
\begin{align}
  r_{ji}=&\frac{\varrho_{ji}}{\sumiM \varrho_{ji}}\\
  \ln \varrho_{ji}=&
\med\Exp{\lndet{\Wmat}}-\frac{d}{2}\ln(2\pi)
  -\med\Exp{\mahP{\phivec_{j}}{\Vtildemat\ytildevec_i}{\Wmat}}
  +\Exp{\ln\pi_{\theta_i}}\;.
\end{align}

The optimum for $\qopt{\pi_\theta}$:
\begin{align}
\qopt{\pi_\theta}=&\mathrm{Dir}(\pi_\theta|\tau)=
C(\tau)\prodiM \pi_{\theta_i}^{\tau_i-1}
\end{align}
where
\begin{align}
  \tau_i=&\Exp{N_i}+\tau_0\\
  C(\tau)=&\frac{\gammaf{\sumiM\tau_i}}{\prodiM\gammaf{\tau_i}}\;
\end{align}

To compute the optimum for $\qopt{\vtildevec'_r}$, we, again,
introduce the parameter $\eta$ to control the weight of the supervised
dataset.
\begin{align}
  \lnqopt{\vtildevec'_r}=&
  \Expcond{\lnProb{\Phimat,\Phimatd,\Ymat,\Ymatd,\theta,\pi_\theta,
      \muvec,\Vmat,\Wmat,\alphavec}}
  {\Ymat,\Ymatd,\theta,\pi_\theta,\Wmat,\alphavec,\vtildevec'_{s\neq r}}
  +\const\\
  =&\Expcond{\lnProb{\Phimat|\Ymat,\theta,\muvec,\Vmat,\Wmat}}
  {\Ymat,\theta,\Wmat,\vtildevec'_{s\neq r}}\nonumber\\
  &+\eta\Expcond{\lnProb{\Phimatd|\Ymatd,\muvec,\Vmat,\Wmat}}
  {\Ymatd,\Wmat,\vtildevec'_{s\neq r}}\nonumber\\
  &+\Expcond{\lnProb{\Vmat|\alphavec}}{\alphavec,\vvec'_{s\neq r}}
  +\Expcond{\lnProb{\muvec}}{\muvec_{s\neq r}}+\const\\
  =&-\med\trace\left(-2\vtildevec'_{r}
    \left(\wbar_{rr}\Cmat_{r}+
      \sum_{s\neq r} \wbar_{rs}
      \left(\Cmat_{s} 
        -\Exp{\vtildevec'_{s}}^{T}\Rmatytilde^{\prime}\right)
      +\beta_r\mutildevec_{0_r}^T\right)\right.\nonumber\\
  &\left.+\vtildevec'_r\vtildevec'^{T}_r
    \left(\diag\left(\alphatbarvec_r\right)
      +\wbar_{rr}\Rmatytilde^{\prime}\right)
  \right)
\end{align}
where $\wbar_{rs}$ is the element $r,s$ of $\Exp{\Wmat}$, 
\begin{align}
  \Cmatytilde=&\sumiM\Exp{\Fvec_i}\Exp{\ytildevec_i}^{T}\\
  \Rmatytilde=&\sumiM \Exp{N_i}\Exp{\scatt{\ytildevec_i}}\\
  \Cmatytilded=&\sumiMd \Fvec_{\rmd_i}\Exp{\ytildevec_{\rmd_i}}^{T}\\
  \Rmatytilded=&\sumiMd N_{\rmd_i}\Exp{\scatt{\ytildevec_{\rmd_i}}}\\
  \Cmatytilde^{\prime}=&\Cmatytilde+\eta\Cmatytilded\\
  \Rmatytilde^{\prime}=&\Rmatytilde+\eta\Rmatytilded\\
  \alphatbarvec_r=&
  \begin{bmatrix}
    \Exp{\alphavec}\\
    \beta_r
  \end{bmatrix}
  \quad\quad 
  \mutildevec_{0_r}=
  \begin{bmatrix}
    \zerovec_{n_y \times 1}\\
    \mu_{0_r}
  \end{bmatrix}
\end{align}
and $\Cmat_r$ is the $r^{th}$ row of $\Cmatytilde^{\prime}$.

Then $\qopt{\vtildevec'_r}$ is a Gaussian distribution:
\begin{align}
  \qopt{\vtildevec'_r}=&
  \Gauss{\vtildevec_{r}'}{\vtbarvec_{r}'}{\iLmatVtr}\\
  \LmatVtr=&\diag\left(\alphatbarvec_r\right)+\wbar_{rr}\Rmatytilde^{\prime}\\
  \vtbarvec_{r}'=&\iLmatVtr\left(\wbar_{rr}\Cmat_{r}^T+
    \sum_{s\neq r} \wbar_{rs}
    \left(\Cmat_{s}^T -\Rmatytilde^{\prime}\vtbarvec_{s}'\right)
    +\beta_r\mutildevec_{0_r}\right)
\end{align}

The optimum for $\qopt{\alphavec}$:
\begin{align}
  \lnqopt{\alphavec}=&
  \Expcond{\lnProb{\Phimat,\Phimatd,\Ymat,\Ymatd,\theta,\pi_\theta,
      \muvec,\Vmat,\Wmat,\alphavec}}
  {\Ymat,\Ymatd,\theta,\pi_\theta,\muvec,\Vmat,\Wmat}+\const\\
  =&\Expcond{\lnProb{\Vmat|\alphavec}}{\Vmat}+\lnProb{\alphavec|a_\alpha,b_\alpha}
  +\const\\
  =& \sum_{q=1}^{n_y} \left(\frac{d}{2}+a_\alpha-1\right)\ln \alpha_q 
  -\alpha_q\left(b_{\alpha}+\med\Exp{\vvec_q^T\vvec_q}\right) +\const\\
\end{align}
Then $\qopt{\alphavec}$ is a product of Gammas:
\begin{align}
  \label{eq:baysplda_v1_apost}
  \qopt{\alphavec}=&\prod_{q=1}^{n_y}
  \Gammad{\alpha_q}{a'_{\alpha}}{b_{\alpha_q}'}\\
  a_{\alpha}'=&a_{\alpha}+\frac{d}{2}\\
  b_{\alpha_q}'=&b_{\alpha}+\med\Exp{\vvec_q^T\vvec_q}
\end{align}

The optimum for $\qopt{\Wmat}$:
\begin{align}
  \lnqopt{\Wmat}=&
  \Expcond{\lnProb{\Phimat,\Phimatd,\Ymat,\Ymatd,\theta,\pi_\theta,
      \muvec,\Vmat,\Wmat,\alphavec}}
  {\Ymat,\Ymatd,\theta,\pi_\theta,\muvec,\Vmat,\alphavec}
  +\const\\
  =&\Expcond{\lnProb{\Phimat|\Ymat,\theta,\muvec,\Vmat,\Wmat}}
  {\Ymat,\theta,\muvec,\Vmat}
  +\eta\Expcond{\lnProb{\Phimatd|\Ymatd,\muvec,\Vmat,\Wmat}}
  {\Ymatd,\muvec,\Vmat}
  +\lnProb{\Wmat}+\const\\
  =&\frac{N^\prime}{2}\lndet{\Wmat}-\frac{d+1}{2}\lndet{\Wmat}
  -\med\trace\left(\Wmat\Kmat\right)+\const
\end{align}
where
\begin{align}
  N^\prime=\Exp{N}+\eta N_{\rmd}\\
  \Kmat=&\Exp{\Smat}+\eta \Smat_{\rmd}
  -\Cmatytilde^{\prime}\Exp{\Vtildemat}^T
  -\Exp{\Vtildemat}\Cmatytilde^T
  +\Expcond{\Vtildemat\Rmatytilde^{\prime}\Vtildemat^T}{\Vtildemat}
\end{align}
Then $\qopt{\Wmat}$ is Wishart distributed:
\begin{align}
  \qopt{\Wmat}=\Wishart{\Wmat}{\Kmat^{-1}}{N^\prime} \quad \textrm{if $N^\prime>d$} \;.
\end{align}

Finally, we evaluate the expectations:
\begin{align}
  \Exp{\yvec_i}=&\ybarvec_i\\
  \Exp{\scatt{\yvec_i}}=&\iLmatyi+\scatt{\ybarvec_i}\\
  \Exp{\scatt{\ytildevec_i}}=&
  \begin{bmatrix}
    \Exp{\scatt{\yvec_i}} & \Exp{\yvec_i} \\
    \Exp{\yvec_i}^T & 1 
  \end{bmatrix}\\
  \Exp{\theta_{ji}}=&r_{ji}\\
  \Exp{\pi_{\theta_i}}=&\frac{\tau_i}{\sumiM\tau_i}\\
  \Exp{\ln \pi_{\theta_i}}=&\psi\left(\tau_i\right)
  -\psi\left(\sumiM\tau_i\right)\\
  \Exp{\alpha_q}=&\frac{a_{\alpha}^\prime}{b_{\alpha_q}^\prime}\\
  \Vtbarmat=&\Exp{\Vtildemat}=
  \begin{bmatrix}
    \vtbarvec_{1}'^T\\
    \vtbarvec_{2}'^T\\
    \vdots\\
    \vtbarvec_{d}'^T
  \end{bmatrix}\\
  \Wbarmat=&\Exp{\Wmat}=N^{\prime}\Kmat^{-1}
\end{align}
\begin{align}
  \Exp{\vvec_q^T\vvec_q}=&\sumrd\Exp{\vvec_{rq}'^T\vvec_{rq}'}\\
  =&\sumrd\iLmat_{\Vtildemat_{r qq}}+\vbarvec_{rq}'^2\\
  \Exp{\Vmat^T\Wmat\Vmat}=&\Exp{\Vmat'\Wbarmat\Vmat'^T}\\
  =&\sumrd\sumsd \wbar_{rs} \Exp{\vvec'_r\vvec'^T_s} \\
  =&\sumrd \wbar_{rr} \SigmatVr + \Vbarmat^T\Wbarmat\Vbarmat\\
  \Exp{\Vmat^T\Wmat\muvec}=&
  \sumrd \wbar_{rr} \SigmatVmur+\Vbarmat^T\Wbarmat\mubarvec\\
  \Exp{\Vtildemat^T\Wmat\Vtildemat}=&
  \sumrd \wbar_{rr} \SigmatVtr + \Vtbarmat^T\Wbarmat\Vtbarmat\\
  \Exp{\mahP{\phivec_{j}}{\Vtildemat\ytildevec_i}{\Wmat}}=&
  \phivec_j^T\Wbarmat\phivec_j
  -2\phivec_j^T\Wbarmat\Vtbarmat\ytbarvec_i
  +\trace\left(\Exp{\Vtildemat^T\Wmat\Vtildemat}
    \Exp{\scatt{\ytildevec_i}}\right)\\
  \Exp{\Vtildemat\Rmatytilde^{\prime}\Vtildemat^T}=&
  \sum_{r=1}^{n_y}\sum_{s=1}^{n_y}
  \rytilders\Exp{\vtildevec_r\vtildevec_s^T}\\
  =&\Vtbarmat\Rmatytilde^{\prime}\Vtbarmat^T+\diag\left(\rhovec\right)
\end{align}
where 
\begin{align}
  \SigmatVtr=&
  \begin{bmatrix}
    \SigmatVr & \SigmatVmur\\
    \SigmatVmur^T & \Sigmatmur
  \end{bmatrix}
  =
  \begin{bmatrix}
    \sigma_{\Vtildemat_{r_{ij}}}
  \end{bmatrix}_{n_y \times n_y}
  =\iLmatVtr\\
  \rhovec=&
  \begin{bmatrix}
    \rho_1 & \rho_2 & \hdots & \rho_d
  \end{bmatrix}^T\\
  \rho_i=&\sum_{r=1}^{n_y}\sum_{s=1}^{n_y}
  \left(\Rmatytilde^{\prime}\circ\iLmatVti\right)_{rs}
\end{align}
and $\circ$ is the Hadamard product. 

\subsubsection{Distributions with deterministic annealing}

If we use annealing, for a parameter $\kappa$, we have
\begin{align}
   \qopt{\Ymat,\Ymatd}=&\prodiM \Gauss{\yvec_i}{\ybarvec_i}{1/\kappa\;\iLmatyi}
   \prodiMd \Gauss{\yvec_{\rmd_i}}{\ybarvec_{\rmd_i}}{1/\kappa\;
     \iLmatydi}\;
\end{align}
\begin{align}
\qopt{\theta}=&\prodjN\prodiM r_{ji}^{\theta_{ji}}
\end{align}
where
\begin{align}
  r_{ji}=&\frac{\varrho_{ji}^\kappa}{\sumiM \varrho_{ji}^\kappa}\;
\end{align}
\begin{align}
\qopt{\pi_\theta}=&\mathrm{Dir}(\pi_\theta|\tau)=
C(\tau)\prodiM \pi_{\theta_i}^{\tau_i-1}
\end{align}
where
\begin{align}
  \tau_i=&\kappa(\Exp{N_i}+\tau_0-1)+1
\end{align}

\begin{align}
  \qopt{\vtildevec'_r}=&
  \Gauss{\vtildevec_{r}'}{\vtbarvec_{r}'}{1/\kappa\;\iLmatVtr}\\
  \qopt{\Wmat}=&\Wishart{\Wmat}{1/\kappa\;\Kmat^{-1}}{\kappa(N^\prime-d-1)+d+1}
  \quad \textrm{if $\kappa(N^\prime-d-1)+1>0$}  \\
  \qopt{\alphavec}=&\prod_{q=1}^{n_y}
  \Gammad{\alpha_q}{a'_{\alpha}}{b_{\alpha_q}'}\\
  a_{\alpha}'=&\kappa\left(a_{\alpha}+\frac{d}{2}-1\right)+1\\
  b_{\alpha_q}'=&\kappa\left(b_{\alpha}+\med\Exp{\vvec_q^T\vvec_q}\right)\;.
\end{align}

\subsection{Variational lower bound}
\label{sec:unsupsplda_v2_lb}

The lower bound is given by:
\begin{align}
  \lowb=&\Expcond{\lnProb{\Phimat|\Ymat,\theta,\muvec,\Vmat,\Wmat}}
  {\Ymat,\theta,\muvec,\Vmat,\Wmat}
  +\Expcond{\lnProb{\Ymat}}{\Ymat} \nonumber\\
  &+\Expcond{\lnProb{\theta|\pi_{\theta}}}{\theta,\pi_\theta}
  +\Expcond{\lnProb{\pi_{\theta}}}{\pi_{\theta}} \nonumber\\  
  &+\Expcond{\lnProb{\Vmat|\alphavec}}{\Vmat,\alphavec}
  +\Expcond{\lnProb{\alphavec}}{\alphavec}
  +\Expcond{\lnProb{\muvec}}{\muvec}
  +\Expcond{\lnProb{\Wmat}}{\Wmat}
  \nonumber\\
  &+\eta \Expcond{\lnProb{\Phimatd|\Ymatd,\muvec,\Vmat,\Wmat}}
  {\Ymatd,\muvec,\Vmat,\Wmat}
  +\eta \Expcond{\lnProb{\Ymatd}}{\Ymatd} \nonumber\\
  &-\Expcond{\lnq{\Ymat}}{\Ymat}-\Expcond{\lnq{\theta}}{\theta}
  -\Expcond{\lnq{\pi_\theta}}{\pi_\theta} \nonumber\\
  &-\Expcond{\lnq{\Vtildemat}}{\Vtildemat}
  -\Expcond{\lnq{\alphavec}}{\alphavec}
  -\Expcond{\lnq{\Wmat}}{\Wmat}
  -\eta \Expcond{\lnq{\Ymatd}}{\Ymatd}\;.
\end{align}

The term $\Expcond{\lnProb{\Phimat|\Ymat,\theta,\muvec,\Vmat,\Wmat}}
{\Ymat,\theta,\muvec,\Vmat,\Wmat}$:
\begin{align}
  \Expcond{\lnProb{\Phimat|\Ymat,\theta,\muvec,\Vmat,\Wmat}}
  {\Ymat,\theta,\muvec,\Vmat,\Wmat}=&
  \frac{\Exp{N}}{2}\Exp{\lndet{\Wmat}}
  -\frac{\Exp{N}d}{2}\ln(2\pi) \nonumber\\
  &-\med\trace\left(\Wbarmat\left(
      \Exp{\Smat}-2\Cmatytilde\Vtbarmat^T
      +\Exp{\Vtildemat\Rmatytilde\Vtildemat^T}
    \right)\right)\\
  =&\frac{\Exp{N}}{2}\ln\overline{\Wmat}-\frac{\Exp{N}d}{2}\ln(2\pi)
  -\med\trace\left(\Wbarmat\Exp{\Smat}\right) \nonumber\\
  &-\med\trace\left(-2\Vtbarmat^T\Wbarmat\Cmatytilde
    +\Exp{\Vtildemat^T\Wmat\Vtildemat}\Rmatytilde
  \right)
\end{align}
where
\begin{align}
  \ln\overline{\Wmat}=&\Exp{\lndet{\Wmat}}\\
  =&\sumid\psi\left(\frac{N^\prime+1-i}{2}\right)+d\ln2 +\lndet{\Kmat^{-1}}
\end{align}
and $\psi$ is the digamma function.

The term $\Expcond{\lnProb{\Phimatd|\Ymatd,\muvec,\Vmat,\Wmat}}
{\Ymatd,\theta,\muvec,\Vmat,\Wmat}$:
\begin{align}
  \Expcond{\lnProb{\Phimatd|\Ymat,\theta,\muvec,\Vmat,\Wmat}}
  {\Ymat,\theta,\muvec,\Vmat,\Wmat}=&
  \frac{N_\rmd}{2}\Exp{\lndet{\Wmat}}-\frac{N_\rmd d}{2}\ln(2\pi) \nonumber\\
  &-\med\trace\left(\Wbarmat\left(
      \Smat_\rmd-2\Cmatytilded\Vtbarmat^T
      +\Exp{\Vtildemat\Rmatytilded\Vtildemat^T}
    \right)\right)\\
  =&\frac{N_\rmd}{2}\ln\overline{\Wmat}-\frac{N_\rmd d}{2}\ln(2\pi)
  -\med\trace\left(\Wbarmat\Smat_\rmd\right) \nonumber\\
  &-\med\trace\left(-2\Vtbarmat^T\Wbarmat\Cmatytilded
    +\Exp{\Vtildemat^T\Wmat\Vtildemat}\Rmatytilded
  \right)
\end{align}

The term $\Expcond{\lnProb{\Vmat|\alphavec}}{\Vmat,\alphavec}$:
\begin{align}
  \Expcond{\lnProb{\Vmat|\alphavec}}{\Vmat,\alphavec}=&
  -\frac{n_y d}{2}\ln(2\pi)+\frac{d}{2}\sum_{q=1}^{n_y}\Exp{\ln\alpha_q}
  -\med\sum_{q=1}^{n_y} \Exp{\alpha_q}\Exp{\vvec_q^T\vvec_q}
\end{align}
where
\begin{align}
  \Exp{\ln\alpha_q}=\psi(a_\alpha^\prime)-\ln b_{\alpha_q}^\prime \;.
\end{align}

The term $\Expcond{\lnProb{\alphavec}}{\alphavec}$:
\begin{align}
  \Expcond{\lnProb{\alphavec}}{\alphavec}=&
  n_y \left(a_\alpha\ln b_\alpha-\ln\gammaf{a_\alpha}\right)+
  (a_\alpha-1)\sum_{q=1}^{n_y}\Exp{\ln\alpha_q}-b_\alpha\sum_{q=1}^{n_y}\Exp{\alpha_q}
\end{align}

The term $\Expcond{\lnProb{\muvec}}{\muvec}$:
\begin{align}
  \Expcond{\lnProb{\muvec}}{\muvec}=&
  -\frac{d}{2}\ln(2\pi)+\frac{1}{2}\sumrd\ln\beta_r
  -\med\sumrd\beta_r\left(\Sigmatmur+\Exp{\mu_r}^2
    -2\mu_{0_r}\Exp{\mu_r}+\mu_{0_r}^2\right)
\end{align}

The term $\Expcond{\lnProb{\Wmat}}{\Wmat}$:
\begin{align}
  \Expcond{\lnProb{\Wmat}}{\Wmat}=&-\frac{d+1}{2}\ln\overline{\Wmat}
\end{align}

The term $\Expcond{\lnq{\Vtildemat}}{\Vtildemat}$:
\begin{align}
  \Expcond{\lnq{\Vtildemat}}{\Vtildemat}=&
  -\frac{d(n_y+1)}{2}\left(\ln(2\pi)+1\right)
  +\med\sumrd\lndet{\LmatVtr}
\end{align}

The term $\Expcond{\lnq{\alphavec}}{\alphavec}$:
\begin{align}
  \Expcond{\lnq{\alphavec}}{\alphavec}=&
  -\sum_{q=1}^{n_y}\Entrop{\q{\alpha_q}}\\
  =&n_y\left((a_\alpha^\prime-1)\psi(a_\alpha^\prime)
    -a_\alpha^\prime-\ln\gammaf{a_\alpha^\prime}\right)
  +\sum_{q=1}^{n_y}\ln b_{\alpha_q}^\prime
\end{align}

The term $\Expcond{\lnq{\Wmat}}{\Wmat}$:
\begin{align}
  \label{eq:EntropWishartW}
  \Expcond{\lnq{\Wmat}}{\Wmat}=&-\Entrop{\q{\Wmat}}\\
  =&\ln B\left(\Kmat^{-1},N\right)
  +\frac{N-d-1}{2}\ln\overline{\Wmat}-\frac{Nd}{2}
\end{align}
where
\begin{align}
  B(\Amat,N)&=\frac{1}{2^{Nd/2}\WZNd}\left|\Amat\right|^{-N/2}\\
  \WZNd=&\WishartZNd{N}{d}
\end{align}

The expressions for the terms $\Expcond{\lnProb{\Ymat}}{\Ymat}$,
$\Expcond{\lnProb{\Ymatd}}{\Ymatd}$, 
$\Expcond{\lnProb{\theta|\pi_{\theta}}}{\theta,\pi_\theta}$,
$\Expcond{\lnProb{\pi_{\theta}}}{\pi_{\theta}}$,
$\Expcond{\lnq{\Ymat}}{\Ymat}$,
$\Expcond{\lnq{\Ymatd}}{\Ymatd}$,
$\Expcond{\lnq{\theta}}{\theta}$ and
$\Expcond{\lnq{\pi_\theta}}{\pi_\theta}$ are the same as the ones in
Section \ref{sec:unsupsplda_v1_lb}.

\subsection{Hyperparameter optimisation}
\label{sec:unsupsplda_v2_hyp}

We can set the Hyperparameters
$\left(\tau_0,\muvec_0,\betavec,a_{\alpha},b_{\alpha}\right)$ 
manually or estimate them from the
development data maximising the lower bound. 

$\tau_0$ can be derived as shown in Section~\ref{sec:unsupsplda_v1_hyp}.

we derive for $a_{\alpha}$
\begin{align}
  \frac{\partial\lowb}{\partial a_{\alpha}}=&
  n_y\left(\ln b_{\alpha}-\psi(a_{\alpha})\right)
  +\sum_{q=1}^{n_y}\Exp{\ln\alpha_q}=0 \quad \implies\\
  \psi(a_{\alpha})=&\ln b_{\alpha}+\frac{1}{n_y}\sum_{q=1}^{n_y}\Exp{\ln\alpha_q}
\end{align}

We derive for $b_{\alpha}$:
\begin{align}
  \frac{\partial\lowb}{\partial b_{\alpha}}=&
  \frac{n_y a_{\alpha}}{b}-\sum_{q=1}^{n_y}\Exp{\alpha_q}=\zerovec\quad \implies\\
  b_{\alpha}=&\left( \frac{1}{n_y a_{\alpha}}\sum_{q=1}^{n_y}\Exp{\alpha_q}\right)^{-1}
\end{align}

We solve these equations with the procedure described
in~\cite{Beal2003}. We write
\begin{align}
  \psi(a)=&\ln b+c\\
  b=&\frac{a}{d}
\end{align}
where
\begin{align}
  c=&\frac{1}{n_y}\sum_{q=1}^{n_y}\Exp{\ln\alpha_q}\\
  d=&\frac{1}{n_y}\sum_{q=1}^{n_y}\Exp{\alpha_q}
\end{align}
Then
\begin{align}
  f(a)=\psi(a)-\ln a + \ln d -c=0
\end{align}

We can solve for $a$ using Newton-Rhaphson iterations:
\begin{align}
  a_{new}=&a-\frac{f(a)}{f^\prime(a)}=\\
  =&a\left(1-\frac{\psi(a)-\ln a + \ln d -c}{a\psi^\prime(a)-1}\right)
\end{align}
This algorithm does not assure that $a$ remains positive. We can put a
minimum value for $a$. Alternatively we can solve the equation for
$\tilde{a}$ such as $a=exp(\tilde{a})$.
\begin{align}
  \tilde{a}_{new}=&\tilde{a}-\frac{f(\tilde{a})}{f^\prime(\tilde{a})}=\\
  =&\tilde{a}-\frac{\psi(a)-\ln a + \ln d -c}{\psi^\prime(a)a-1}
\end{align}
Taking exponential in both sides:
\begin{align}
  a_{new}=a\exp\left(-\frac{\psi(a)-\ln a + \ln d -c}{\psi^\prime(a)a-1}\right)
\end{align}

We derive for $\muvec_0$:
\begin{align}
  \frac{\partial\lowb}{\partial \muvec_0}=&\zerovec \quad \implies\\
  \muvec_0=&\Exp{\muvec}
\end{align}

We derive for $\betavec$:
\begin{align}
  \frac{\partial\lowb}{\partial \betavec}=&\zerovec \quad \implies\\
  \betavec_r^{-1}=&\Sigmatmur+\Exp{\mu_r}^2
  -2\mu_{0_r}\Exp{\mu_r}+\mu_{0_r}^2
\end{align}

If we take an isotropic prior for $\muvec$:
\begin{align}
  \betavec^{-1}=&\frac{1}{d}\sumrd \Sigmatmur+\Exp{\mu_r}^2
  -2\mu_{0_r}\Exp{\mu_r}+\mu_{0_r}^2
\end{align}

\subsection{Some ideas}

What we expect from this model is:
\begin{itemize}
  \item We expect that taking into account the full posterior of the
    parameters of the
    SPLDA, we will obtain a better estimation of the labels and the
    number of speakers.
  \item The variances of $\Vmat$ and $\Wmat$ decrease as the number of
    speakers and segments, respectively, grow. Thus, we expect a larger
    improvement in cases where we have scarce adaptation data.
  \item We can analyse, how the labels affect the posteriors of the
    parameters. I have the intuition that if the labels are wrong the
    variance of $\Vmat$ should be larger than if the labels are right.
  \item From $\q{\alpha}$, we can infer the best value for
    $n_y$. If the $\Exp{\alpha_q}$ (prior precision of $\vvec_q$) is
    large, $\vvec_q$ will tend to be small as can be seen in 
    Equation~\eqref{eq:unsupsplda_v2_priors_V}.
\end{itemize}

\bibliographystyle{IEEEbib}
\bibliography{villalba}

\end{document}